\ifdefined\pdfoutput\pdfoutput=1\fi

\documentclass[11pt]{article}

\usepackage[final]{acl}

\usepackage{times}
\usepackage{latexsym}

\usepackage[T1]{fontenc}

\usepackage[utf8]{inputenc}

\usepackage{microtype}

\usepackage{inconsolata}

\usepackage{graphicx}

\usepackage{enumitem}

\usepackage{booktabs}

\usepackage{amsmath}

\usepackage{pifont}
\usepackage[most]{tcolorbox}
\tcbuselibrary{listings,breakable}

\usepackage{tikz}
\usepackage{hyperref}       
\usepackage{url}            
\usepackage{booktabs}       
\usepackage{amsfonts}       
\usepackage{nicefrac}       
\usepackage{microtype}      
\usepackage{xcolor}         
\usepackage{textcomp}
\usepackage{graphicx}
\usepackage{multirow}
\usepackage{color}
\usepackage{threeparttable} 
\usepackage{tablefootnote} 
\usepackage{arydshln} 
\usepackage{amssymb} 
\usepackage{pifont} 
\usepackage{wrapfig}
\usepackage{verbatim}
\usepackage{subscript}
\usepackage{enumitem}
\usepackage{amsmath}
\usepackage{multirow}
\usepackage{colortbl}
\usepackage{booktabs}
\usepackage{setspace}
\usepackage{hhline}
\usepackage{subcaption}
\usepackage{makecell}
\usepackage{mathrsfs}
\usepackage{tikz}
\usepackage{tabularx}
\usepackage{amssymb}
\usepackage{fontawesome5}

\newcommand{\MYNAME}{\textsc{Dragon}}
\newcommand{\MYBench}{\textsc{DragonBench}}

\title{\textcolor{cyan}{\MYNAME}: \textcolor{cyan}{D}omain-specific \textcolor{cyan}{R}obust \textcolor{cyan}{A}utomatic Data \textcolor{cyan}{G}eneration \\for RAG \textcolor{cyan}{O}ptimizatio\textcolor{cyan}{n}}

\author{%
  \textbf{Haiyang Shen$^{1,2,}$\thanks{Email: hyshen@stu.pku.edu.cn},\quad Hang Yan$^{3,}$\thanks{Project Leader: hyan@cuhk.edu.hk},\quad Zhongshi Xing$^{4}$,\quad Mugeng Liu$^{2}$}\\\textbf{Yue Li$^{5}$, Zhiyang Chen$^{1,2}$, \quad Yuxiang Wang$^{2}$, \quad Jiuzheng Wang$^{2}$, \quad Yun Ma$^{1,2,}$}\thanks{Corresponding: mayun@pku.edu.cn}\vspace{8pt}\\
  $^{1}$Institute for Artificial Intelligence, Peking University \\
  $^{2}$School of Computer Science, Peking University \\
  $^{3}$The Chinese University of Hong Kong \\
  $^{4}$School of Computer Science, Sun Yat-sen University \\
  $^{5}$School of Software \& Microelectronics, Peking University
}

\newcommand*\circledwhite[1]{\tikz[baseline=(char.base)]{
            \node[shape=circle,draw,inner sep=0pt, minimum size=9pt] (char) {\scriptsize #1};}}

\newcommand{\BoldBlueUparrow}{%
    \begin{tikzpicture}[baseline=(base)]
        \node (base) at (0,0.05) {};
        \draw[line width=1.pt, blue, ->] (0,0) -- (0,0.3);
    \end{tikzpicture}%
}
\newcommand{\BoldRedDownarrow}{%
    \begin{tikzpicture}[baseline=(base)]
        \node (base) at (0,0.05) {};
        \draw[line width=1.pt, red, ->] (0,0.3) -- (0,0);
    \end{tikzpicture}%
}

\newcommand{\bluetext}[1]{\textcolor{blue}{#1}}

\newcommand{\redtext}[1]{\textcolor{red}{#1}}

\newcommand{\graytext}[1]{\textcolor{gray}{#1}}

\newcommand{\SH}{
    \begin{tikzpicture}[baseline=(S.base)]
        \node[anchor=base, scale=1, text=blue!50!black, font=\bfseries] (S) {\textsc{Sh}};
    \end{tikzpicture}
}
\newcommand{\MH}{
    \begin{tikzpicture}[baseline=(M.base)]
        \node[anchor=base, scale=1, text=black!50!black, font=\bfseries] (M) {\textsc{Mh}};
    \end{tikzpicture}
}

\newcommand{\Ori}{\textsc{Ori}}
\newcommand{\Bef}{\textsc{Bef}}

\newcommand{\SmallOri}{
    \begin{tikzpicture}[baseline=(O.base)]
        \node[anchor=base, scale=0.8, text=blue!50!black, font=\bfseries] (O) {\textsc{Ori}};
    \end{tikzpicture}
}
\newcommand{\SmallBef}{
    \begin{tikzpicture}[baseline=(B.base)]
        \node[anchor=base, scale=0.8, text=blue!50!black, font=\bfseries] (B) {\textsc{Bef}};
    \end{tikzpicture}
}
\newcommand{\SmallAft}{
    \begin{tikzpicture}[baseline=(A.base)]
        \node[anchor=base, scale=0.8, text=black!50!black, font=\bfseries] (A) {\textsc{Aft}};
    \end{tikzpicture}
}

\newcommand{\citeRR}{\cite{he2022rethinkingretrievalfaithfullarge}}
\newcommand{\citeReACT}{\cite{yao2023react}}
\newcommand{\citevanillaRAG}{\cite{gao2024retrievalaugmentedgenerationlargelanguage}}
\newcommand{\citeFlare}{\cite{jiang-etal-2023-active}}

\newcommand{\citeStella}{\cite{zhang2025jasperstelladistillationsota}}
\newcommand{\citeGTE}{\cite{gte_multilingual_base}}
\newcommand{\citeSnowflake}{\cite{merrick2024embeddingclusteringdataimprove}}
\newcommand{\citeRubert}{\cite{rubert_tiny_turbo}}
\newcommand{\citeMedEmbed}{\cite{medembed}}

\begin{document}
\maketitle
\begin{abstract}

Retrieval-augmented generation (RAG) can substantially enhance the performance of LLMs on knowledge-intensive tasks. Various RAG paradigms—including vanilla, planning-based, and iterative RAG—all depend on a robust retriever, yet existing retrievers rely heavily on public knowledge and often falter when faced with domain-specific queries. To address these limitations, we introduce \MYNAME, a framework that combines a data-construction modeling approach with a scalable synthetic data-generation pipeline, specifically designed to optimize domain-specific retrieval performance and bolster retriever robustness. To evaluate RAG performance on domain-specific RAGs, we propose \MYBench, a benchmark spanning 8 domain-specific document collections across 4 distinct fields and featuring a wide spectrum of query complexities, answerability, and hop numbers.
Leveraging \MYNAME, we generate a large-scale synthetic dataset—encompassing both single-hop and multi-hop queries—to enrich retriever training. Extensive experiments demonstrate that retrievers trained on this data yield significant performance gains and exhibit strong cross-domain generalization. Moreover, when our optimized retrievers are integrated into vanilla, planning-based, and iterative RAG paradigms, we observe consistent end-to-end improvements in system accuracy. 

\end{abstract}

\section{Introduction}
\label{sec:Introduction}

Retrieval-Augmented Generation (RAG)~\cite{hu2024rag,zhao2024retrieval,gao2024retrievalaugmentedgenerationlargelanguage} can enhance the performance of the large language models (LLM) on knowledge-intensive tasks. Formally, given user queries \( \mathscr{Q} = \{q_i\}_{i=1}^m \) and document set \( \mathbb{D} = \{D_i\}_{i=1}^N \), the goal of RAG is to give an answer \( \hat{\mathscr{A}}_i \) for each query \( q_i \) by referencing \( \mathbb{D} \). The retriever \( \mathcal{R} \) selects documents \( \mathcal{R}(q_i, \mathbb{D}; \theta) \rightarrow \hat{R}_i \) from \( \mathbb{D} \), and the LLM \( 
\mathcal{G} \) generates an answer, represented as \( \mathcal{G}(q_i, \hat{R}_i; \Theta) \rightarrow \hat{\mathscr{A}}_i \).

Existing RAGs~\cite{hu2024rag} include 3 structures: vanilla, planning-based, and iterative. \textit{In vanilla RAG}~\cite{10.5555/3524938.3525306,DBLP:conf/icml/BorgeaudMHCRM0L22}, the retriever retrieves once, and the generator gives the answer. In planning-based RAG~\cite{he2022rethinkingretrievalfaithfullarge}, a planner first devises a strategy composed of multiple sub-queries, followed by the retriever doing retrievals, and the generator synthesizing. In iterative RAG~\cite{yao2023react,jiang-etal-2023-active}, the LLM interacts iteratively with the retriever, and when the LLM determines that sufficient information has been obtained, it hands over to the generator for synthesis. Regardless of the structure, a robust retriever is crucial for the entire RAG system.

In this paper, we focus on optimizing dense retrievers for domain-specific corpora to boost both retrieval robustness and end-to-end RAG performance. Existing dense retrievers are usually trained on general‐purpose sources (e.g., Wikipedia), which hampers their effectiveness on domain-specific content~\cite{lee2025nvembed}. This shortfall stems from (1) domain-specific terminology and query styles, (2) semantic ambiguities or some specific constraints that misalign queries and documents, and (3) RAG’s reliance on LLMs to decompose queries into sub‐queries of varying completeness, further complicating matching. In specialized domains, authors often omit intermediate assumptions, use diverse and scattered expressions, and require multi-step, cross-paragraph or cross-document reasoning—compounding misalignment and sub-query retrieval challenges. Therefore, a robust retriever must handle a wide range of logical complexities and retrieve documents relevant to both full queries and their partial sub-queries.

In this paper, we aim to enhance the retriever used in RAG from the perspective of data synthesis. We propose \MYNAME, which incorporates an RAG data synthesis model to capture the complex mapping relationships among documents, queries, ground-truth answers, relevant clues, and their connections to source documents, along with a specific implementation. \MYNAME~can be used for large-scale RAG data generation for documents in any specific domain, supporting both single-hop and multi-hop queries. It enables the simulation of query generation with varying logical complexities and clue completeness through paraphrasing. Using \MYNAME, we synthesize a diverse dataset to enhance the retriever's performance and the end-to-end performance of the entire RAG system.

To provide the \MYNAME~ with various domain document corpora for synthesizing data and constructing a benchmark to evaluate the domain-specific retriever performance, we collect 8 domain-specific document collections across 4 domains and build \MYBench. \MYBench~includes queries with varying hop counts. The query may be fully or partially addressed by certain sentences in the corresponding corpus, reflecting situations that frequently occur in real life. It also implements sentence-level citations for the clues supporting the answers and provides detailed evaluation criteria. Building on these evaluation criteria, we developed the criteria-based score generation (CSG) to assess the fidelity of the generator. We find that CSG offers greater stability compared to the basic LLM-as-a-Judge approach~\cite{gu2025surveyllmasajudge}, same as RocketEval~\cite{wei2025rocketeval} and BiGGenBench~\cite{kim-etal-2025-biggen} show. 

To validate the improvement of \MYNAME~on retrievers, we select 6 top retrievers from the MTeb leaderboard~\cite{muennighoff-etal-2023-mteb} with varying sizes (33\textasciitilde~611M) and context lengths (512\textasciitilde~8192). Using contrastive learning~\cite{izacard2022unsupervised} and ANCE hard negative sampling~\cite{xiong2021approximate}, we demonstrate significant performance improvement on all 4 large-scale document collections~\cite{hearthstone_wiki,zelda_wiki,drugs_com,mayo_clinic}. The ablation study reveals that retriever gains stem not only from in-domain training but also critically from complex logical reformulations and clue-completeness constructions, which substantially boost performance and robustness. Moreover, these enhancements can also generalize to out-of-domain data: a retriever trained on the Zelda~\cite{zelda_wiki} corpus consistently delivers significant improvements on four other domain‐specific datasets~\cite{stanford_admissions,berkeley_admissions,cyotek,notion_help}. Finally, when integrated into various RAG paradigms, our optimized retriever yields stable end-to-end performance gains across the entire system.

In summary, this paper makes the following key contributions:
\begin{itemize}[left=0.1cm]
\setlength\itemsep{0.1em}
\item We propose \MYNAME, a data synthesis method, along with an implementation. \MYNAME~formally constructs mapping relationships among documents, queries, answers, relevant clues, and their connections to source documents. Based on \MYNAME, we introduce a large-scale synthetic dataset characterized by varying logical complexities, clue completeness, and hop counts.
\item We introduce \MYBench, consisting of 8 domain-specific datasets across 4 domains. \MYBench~provides extensive domain coverage, diverse hop coverage, varying levels of question answerability, and finer citation granularity. Utilizing its per-question scoring criteria, we also propose the CSG score, demonstrating its stability over the LLM-as-a-Judge metric.
\item Extensive experiments validate \MYNAME's improvements in the retriever across various domains. While existing work primarily focuses on optimizing LLMs themselves, we demonstrate that optimizing the retriever with synthetic data can also enhance the end-to-end consistency performance of the entire RAG system.
\end{itemize}
\section{Related Work}
\label{sec:relatedwork}

In this Section, we first discuss existing work for RAG (\ref{sec:Retrieval-augmented-Generation}). Second, we introduce a series of optimization approaches utilizing generated data (\ref{sec:Data-Synthesis-Optimization-Methods}), particularly related to RAG. Finally, we review existing RAG benchmarks and highlight the distinctions between them and \MYBench~(\ref{sec:RAG-Benchmarks}).

\subsection{Retrieval-Augmented Generation}
\label{sec:Retrieval-augmented-Generation}

RAG enhances the understanding of queries by incorporating external knowledge. In vanilla RAG~\cite{10.5555/3524938.3525306,DBLP:conf/icml/BorgeaudMHCRM0L22}, the generator and retriever interact once, relying entirely on the retriever's capabilities. Because the retriever's intelligence is limited, the RAG's understanding of user queries depends largely on the retriever~\cite{zhao2024retrieval,hu2024rag}. Consequently, recent studies~\cite{he2022rethinkingretrievalfaithfullarge,yao2023react,jiang-etal-2023-active,pang2024empowering} emphasize leveraging LLMs to enhance query interpretation. Notable approaches include planning~\cite{he2022rethinkingretrievalfaithfullarge} and iterative RAG~\cite{yao2023react,jiang-etal-2023-active}.

Regardless of the structure, a robust retriever is crucial for the entire RAG system. The performance of the RAG system depends not only on the LLM's ability to paraphrase, decompose, and plan but also heavily relies on the retriever's adaptability to varying question complexities, answerability, and domain-specific styles.

\subsection{Data Synthesis Optimization Methods}
\label{sec:Data-Synthesis-Optimization-Methods}

The huge success of scaling law~\cite{kaplan2020scalinglawsneurallanguage} in LLM has spurred data synthesis methods. In NLP~\cite{deepseekai2025deepseekr1incentivizingreasoningcapability}, model distillation from larger LLMs can be used to enhance the capabilities of smaller LLMs. In LLM-based agents~\cite{qin2024toolllm}, the integration of LLMs with tools can be optimized using synthetic code~\cite{zhang2024xlamfamilylargeaction} and workflow~\cite{shen2025shortcutsbench,zeng2023agenttuningenablinggeneralizedagent,yin2025magnetmultiturntoolusedata}. In RAG, data synthesis methods encompass strategies for both the retriever~\cite{ma2025dramadiverseaugmentationlarge,kim-baek-2025-syntriever,zhou2024megapairsmassivedatasynthesis} and the LLM~\cite{devine2025aloftragautomaticlocalfine,jadon2025enhancingdomainspecificretrievalaugmentedgeneration,wei2025instructraginstructingretrievalaugmentedgeneration,chen-etal-2024-shot}. For the retriever, data synthesis focuses on generating query-positive-negative triplets using LLMs.

However, existing work still lacks a conceptually clear, domain-specific paradigm for automated data generation and augmentation. In particular, there is a lack of robust studies on enhancing retrievers with respect to the logical complexity and answer completeness of questions, as well as research on their impact on the entire RAG system.

\subsection{RAG Benchmarks}
\label{sec:RAG-Benchmarks}

RAG benchmarks can be categorized into single, multi-hop, and various hybrid types. These datasets are primarily derived from sources such as Wikipedia, e.g., Natural Questions~\cite{kwiatkowski-etal-2019-natural}, TriviaQA~\cite{joshi-etal-2017-triviaqa}, SQuAD~\cite{rajpurkar-etal-2016-squad}, PopQA~\cite{mallen-etal-2023-trust}, HotpotQA$^{*}$~\cite{yang-etal-2018-hotpotqa}, 2WikiMultiHopQA$^{*}$~\cite{ho-etal-2020-constructing}, R4C$^{*}$~\cite{inoue-etal-2020-r4c}, MuSiQue$^{*}$~\cite{trivedi-etal-2022-musique}, the Freebase knowledge base, e.g., WebQuestions~\cite{berant-etal-2013-semantic}, or Bing search logs, e.g., MS MARCO~\cite{DBLP:conf/nips/NguyenRSGTMD16}. Hybrid types such as CRAG~\cite{yang2024crag} and RAGBench~\cite{friel2025ragbenchexplainablebenchmarkretrievalaugmented} include a relatively rich set of rag data.

Compared with them, \MYBench~covers 8 domain-specific datasets across 4 areas, featuring varying logical complexity, query answerability, and hops. Additionally, we provided a fine-grained scoring rubric for each question to mitigate the instability issues associated with the naive use of LLM-as-a-Judge for evaluation.
\section{\MYNAME}
\label{sec:method}

\subsection{Modeling}
\label{sec:modeling}

\paragraph{Design Rationale.}
We aim to make the data synthesis broadly applicable across heterogeneous domain corpora (documents, web pages, tables).
(1) \textit{Entity graph for multi-hop:} We adopt an entity-centric graph to capture domain-specific terminology, style, and cross-document links in a format-agnostic way. This provides a general mechanism to support single-hop and multi-hop question construction by explicitly bridging sentences across documents via entities and relations.
(2) \textit{Rephrasing rules for diversity:} We intentionally keep rephrasing rules simple and randomly initialized to encourage a diverse set of equivalence-preserving transformations. The goal is to cover more domain-specific language phenomena (e.g., terminology, stylistic variation) and maximize query diversities (e.g., metaphor, constraint inclusion).

To unify single and multi-hop, we introduce an intermediate factor to connect different documents, which we refer to as clues. Clues serve as the basis for answering queries and are derived from sentences in one or more documents. For multi-hop data, clues act as bridges linking sentences across documents; for single-hop data, clues correspond to sentences within a single document.  
Our goal is to construct synthetic data: \(\mathscr{G} = (\mathscr{D}, \mathscr{Q}, \mathscr{C}, \mathscr{A}, \mathscr{M})\). The dataset \(\mathscr{G}\) consists of:

\noindent\textbf{Document Set (\(\mathscr{D}\))} A collection of docs \(\mathscr{D} = \{D_i\}_{i=1}^M\), where each \(D_i\) is selected from the subset \(\mathscr{D} \subseteq \mathbb{D}\) utilized by \MYNAME~to synthesize data. \(\mathscr{D}\) is input and all others are output.

\noindent\textbf{Query Set (\(\mathscr{Q}\))} A set of queries \(\mathscr{Q} = \{q_i\}_{i=1}^m\), each associated with a answer \(\mathscr{A}_i\) in \(\mathscr{A}\).

\noindent\textbf{Clue Set (\(\mathscr{C}\))} The set of clues required to answer each query \(q_i\), denoted as \(\mathscr{C} = \{\mathscr{C}_i\}_{i=1}^m\). Each clue set \(\mathscr{C}_i = \{c_{i,j}\}_{j=1}^{u_i}\) consists of clues derived from sentences in documents. These clues serve as intermediate links that connect different documents for addressing multi-hop questions, facilitating query construction.

\noindent\textbf{Answer Set (\(\mathscr{A}\))} \(\mathscr{A} = \{ \mathscr{A}^o_i, \mathscr{A}^{v}_{i,j} \}_{i=1,j=1}^{m,P_i}\). For convenience, we use \( \mathscr{A}_{i} \) to uniformly represent both \( \mathscr{A}^{o}_{i} \) and \( \mathscr{A}^{v}_{i} \) with specific combination of retrieved documents.

\begin{itemize}[left=0cm]
\setlength\itemsep{0.1em}
    \item \(\mathscr{A}^o = \{\mathscr{A}^o_i\}_{i=1}^m\) comprises standard answers, representing the complete response when all necessary documents are retrieved. Each \(\mathscr{A}^o_i\) consists of multiple sentences, \(\mathscr{A}^o_i = \{s_{i,j}\}_{j=1}^{n_i}\), where each sentence may include a citation to the original text, as shown in Equation \ref{equ:M2}: \((D_{i,j,k}, S_{i,j,k,l})\). In the actual data generation, we can first add a reference to the clue after each \(s_{i,j}\) as \( \mathscr{M}_{1} \) shows, and then map the clue \( \mathscr{C}_{i} \) to get \((D_{i,j,k}, S_{i,j,k,l})\) through \( \mathscr{M}_{2} \).
    
    \item For each \(\mathscr{A}^o_i\), there exists a set of answer variants \(\{\mathscr{A}^{v}_{i,j}\}_{j=1}^{P_i}\), where \(P_i = \sum_{k=1}^{p_i} \binom{k}{p_i}\) and \(p_i\) is the number of documents (e.g., hops). These variants \(\mathscr{A}^{v}_{i,j}\) capture alternative forms of \(\mathscr{A}^o_i\) when one or more supporting documents are missing from retrieval. 
\end{itemize}

\noindent\textbf{Mapping Relationships (\(\mathscr{M}\))}
\begin{itemize}[left=0cm]
\setlength\itemsep{0.1em}
    \item \textbf{Answer-to-Clue Mapping (\(\mathscr{M}_1\))} This mapping associates sentence \(s_{i,j}\) in the answer \(\mathscr{A}_i\) with one or more corresponding clues.
    
    \item \textbf{Clue-to-Document Mapping (\(\mathscr{M}_2\))} This mapping links each clue \(c_{i,j}\) to the specific document-sentence pairs that support it:
    \begin{equation}
    \label{equ:M2}
    \begin{aligned}
        \mathscr{M}_2^{D}(c_{i,j}) &= \left\{ D_{i,j,k} \right\}_{k=1}^{r_{i,j}}, \\
        \mathscr{M}_2^{S}(c_{i,j}) &= \left\{ \left\{ (D_{i,j,k}, S_{i,j,k,l}) \right\}_{l=1}^{y_{i,j,k}} \right\}_{k=1}^{r_{i,j}}
    \end{aligned}
    \end{equation}
    where \(D_{i,j,k}\) is the \( k\text{-th} \) document associated with clue \(c_{i,j}\), \(S_{i,j,k,l}\) is the specific sentence within \(D_{i,j,k}\) that provides the information to answer query \(q_i\), \(y_{i,j,k}\) is the number of key sentences required from \(D_{i,j,k}\) to answer \(q_i\), and \(r_{i,j}\) is the number of associated documents with clue \( c_{i,j} \).
\end{itemize}

\vspace{0.5em}

Using the mappings \(\mathscr{M}_1\) and \(\mathscr{M}_2\), for each query \(q_i \in \mathscr{Q}\), we can identify the corresponding document set \(\mathcal{D}(q_{i})\) and the sentences \(\mathcal{S}(q_{i})\) required to answer the query. Specifically, the answer \(\mathscr{A}_i\) for \(q_i\) is first mapped to its set of clues \(\mathscr{C}_i\) via the answer-to-clue mapping \( \mathscr{M}_1(\mathscr{A}_i) \).
Subsequently, the document set \(\mathcal{D}(q_{i})\) is defined as the union of all documents associated with each clue \(c_{i,j} \in \mathscr{C}_i\) through the clue-to-document mapping \(\mathscr{M}_2\):
\begin{equation}
\begin{split}
\mathcal{D}(q_{i})
&= \bigcup_{c_{i,j}\in\mathscr{M}_1(\mathscr{A}_i)}
   \{\,D\mid D\in\mathscr{M}^{D}_2(c_{i,j})\}
\end{split}
\end{equation}

\noindent where \(|\mathcal{D}(q_{i})| = p_i\). \(p_i\) denotes the number of required documents. The set of specific sentences \(\mathcal{S}(q_{i})\) required to answer \(q_i\) is obtained by aggregating all supporting sentences from the mapped documents for each clue in \(\mathscr{C}_i\):

\vspace{-1em}
\begin{align}
  \mathcal{S}(q_{i})
    &= \bigcup_{c_{i,j}\in\mathscr{M}_1(\mathscr{A}_i)}
       \label{equ:M3} \\[-1ex]
    &\quad\{\,S_{i,j,k,l}\mid (D_{i,j,k},\,S_{i,j,k,l})
       \in \mathscr{M}^{S}_2(c)\,\}
       \nonumber
\end{align}
\vspace{-1em}

\noindent where \(|\mathcal{S}(q_{i})| = w_i\). \(w_i\) represents the total supporting sentences for \(q_i\). To optimize the RAG system, both the retriever and generator components are trained using the synthetic dataset \(\mathscr{G}\).

For the retriever, the required synthetic data consists of \( \{ \mathscr{D}, \mathscr{Q} \} \), and the mapping from \( \mathscr{Q} \) to \( \mathscr{D} \) derived from \( \mathscr{M} \). For the entire RAG, the complete structure \( \mathscr{G} = (\mathscr{D}, \mathscr{Q}, \mathscr{C}, \mathscr{A}, \mathscr{M}) \) is necessary. Additionally, it is essential to construct actual retrieval results using a real retriever based on \( \mathscr{G} \), as the document set provided to the generator during real RAG operation may not fully answer the query. Therefore, we need to consider generating corresponding \( \mathscr{A}^{v}_{i,j} \) under various retrieval scenarios.

\subsection{Implementation}
\label{sec:Implementation}

\begin{figure*}[ht!]
  \centering
  \scalebox{1.0}[1]{%
    \includegraphics[width=\linewidth]{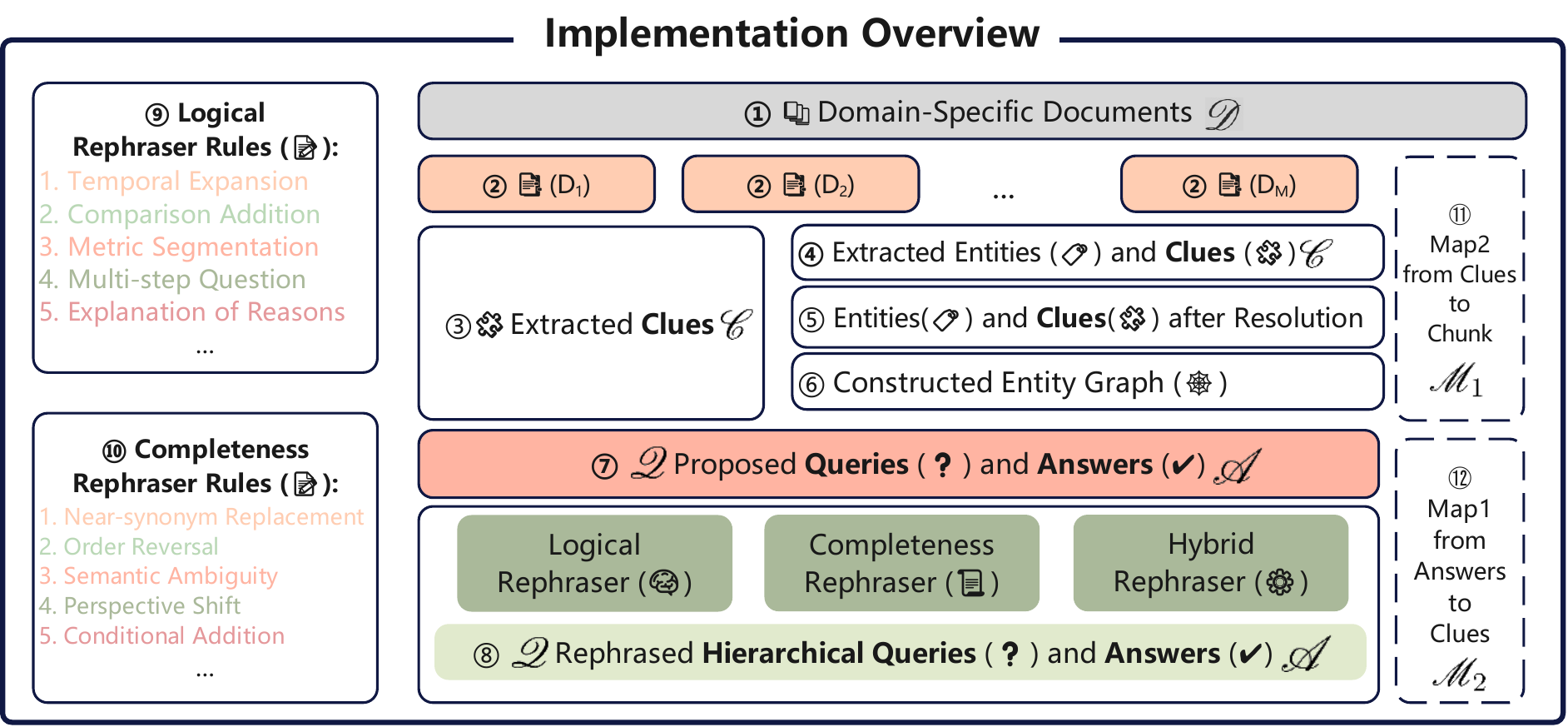}%
  }
  \caption{A specific implementation of the \MYNAME.}
  \label{fig:AutomaticDataConstructionPipe}
  \vspace{-0.5em}
\end{figure*}

Figure \ref{fig:AutomaticDataConstructionPipe} illustrates an implementation of \MYNAME. The process begins by partitioning documents \textbf{(\circledwhite{1})} into multiple chunks, thereby forming the document set \(\mathscr{D}\) \textbf{(\circledwhite{2})}. We chunk documents to align retrieval granularity with the retriever context length. From \(\mathscr{D}\), factual clues are extracted. For single-hop queries, clues are directly extracted \textbf{(\circledwhite{3})}. In the case of multi-hop queries, connections between multiple documents are established by entities extracted from clues to facilitate question construction \textbf{(\circledwhite{4})}. After performing entity resolution \textbf{(\circledwhite{5})}, an entity graph is constructed \textbf{(\circledwhite{6})}. Subsequently, the extracted clues are input into an LLM to generate straightforward questions \textbf{(\circledwhite{7})}. For multi-hop questions, multiple entity-centric clues are provided to the LLM to generate more complex, entity-related queries \textbf{(\circledwhite{7})}. Following this, based on predefined equivalence transformation rules \textbf{(\circledwhite{9})}, the LLM is prompted to iteratively reformulate the generated questions \textbf{(\circledwhite{8})} and vary the completeness of the clues \textbf{(\circledwhite{10})}. This results in the construction of data with multiple levels of comprehension difficulty and varying degrees of clue completeness.

\setlength{\tabcolsep}{6pt}

\begin{table*}
\centering
\caption{Basic information of \MYBench. The Games and Mediguides corpus is relatively large, making it suitable for synthetic data, while the Universities and Documentation corpus is comparatively small, ideal for validating cross-domain generalization.}
\vspace{-0.5em}
\label{tab:Domain-specific-Documents}
\renewcommand{\arraystretch}{1.1}
\scalebox{0.9}{
\begin{tabular}{ccc|cc|cc|cc}
\hline
\multirow{2}{*}{\textbf{\MYBench}} & \multicolumn{2}{c|}{\textbf{Games}} & \multicolumn{2}{c|}{\textbf{Mediguides}} & \multicolumn{2}{c|}{\textbf{Universities}} & \multicolumn{2}{c}{\textbf{Documentation}} \\ \cline{2-9} 
 & \multicolumn{1}{c|}{\textbf{Hearthstone}} & \textbf{~Zelda} & \multicolumn{1}{c|}{\textbf{Drug}} & \textbf{Mayoclinic} & \multicolumn{1}{c|}{\textbf{Stanford}} & \textbf{Berkeley} & \multicolumn{1}{c|}{\textbf{Cyotek}} & \textbf{Notion} \\ \hline
\textbf{\# Docs} & \multicolumn{1}{c|}{15695} & 10665 & \multicolumn{1}{c|}{77348} & 2542 & \multicolumn{1}{c|}{65} & 280 & \multicolumn{1}{c|}{808} & 475 \\
\textbf{\# Avg Tokens} & \multicolumn{1}{c|}{1112} & 746 & \multicolumn{1}{c|}{2159} & 1369 & \multicolumn{1}{c|}{1052} & 1283 & \multicolumn{1}{c|}{478} & 4382 \\ \hline
\textbf{\# Single Hop} & \multicolumn{1}{c|}{873} & 870 & \multicolumn{1}{c|}{826} & 862 & \multicolumn{1}{c|}{272} & 860 & \multicolumn{1}{c|}{653} & 877 \\
\textbf{\# Multi Hop} & \multicolumn{1}{c|}{162} & 390 & \multicolumn{1}{c|}{183} & 234 & \multicolumn{1}{c|}{162} & 398 & \multicolumn{1}{c|}{259} & 447 \\ \hline
\end{tabular}
}
\end{table*}
\section{Document Corpus and \MYBench}
\label{sec:Documents-and-MYBench}

To provide the \MYNAME~ with a corpus for synthesizing and constructing a benchmark, we curate 8 domain-specific document corpora across 4 domains and build \MYBench~(\ref{sec:Domain-specific-Documents}). We introduce the construction process (\ref{sec:MYBench-Construction}). Then we introduce the evaluation metrics to assess the performance of the retriever and the generator (\ref{sec:TaskMetrics}).

\subsection{Domain-specific Corpus}
\label{sec:Domain-specific-Documents}

The principle for data selection was to cover a broad range of applications with high practical value. Thus, we chose datasets from the following areas: (1) game guides (Hearthstone~\cite{hearthstone_wiki} and Zelda~\cite{zelda_wiki}); (2) medical guides (Drug~\cite{drugs_com} and Mayoclinic~\cite{mayo_clinic}); (3) university admissions (Stanford~\cite{stanford_admissions} and Berkeley~\cite{berkeley_admissions}); and (4) software documentation (Cyotek~\cite{cyotek} and Notion~\cite{notion_help}). These datasets were thoroughly cleaned using an LLM. For detailed cleaning procedures, please refer to Appendix~\ref{app:cleaning}. Our corpus selection follows two principles: \emph{diversity across domains} and \emph{public accessibility} of source documents. Details on annotator qualifications and selection are provided in Appendix~\ref{app:experts}.

\subsection{\MYBench~Construction}
\label{sec:MYBench-Construction}

We first selected sub-documents and then segmented these documents into chunks. We then built an entity–relationship graph from these chunks and, based on its connections, assigned related chunks to domain experts (e.g., Zelda and Hearthstone enthusiasts) to generate single‐hop and multi‐hop datasets. Once questions were generated and answered, we asked annotators to reformulate them according to their personal preferences, thereby producing a rich collection of questions, answers, and their source documents and sentences. Building on this, we added question‐quality assessment and answer‐verification steps to filter the items. Finally, we used an LLM to draft scoring rubrics for each question and had the original question authors conduct two rounds of review—only questions whose rubrics passed both reviews were retained. We finally get \MYBench. Importantly, the construction of the evaluation set is independent of our synthesized training data, which helps ensure fair assessment of retriever and RAG performance.
Annotator recruitment criteria and qualifications are summarized in Appendix~\ref{app:experts}. The final dataset is presented in Table~\ref{tab:Domain-specific-Documents}. For detailed construction procedures and examples, please refer to the Appendix~\ref{app:bench}.

\setlength{\tabcolsep}{6pt}

\begin{table}
\centering
\caption{The score inconsistency between different model on Zelda. The consistency of CSG between models and humans significantly outperforms the LJ}
\label{tab:the_superiority_of_CSG_scores}
\renewcommand{\arraystretch}{1.1}
\scalebox{0.8}{
\begin{tabular}{crr}
\hline
                                                         & \multicolumn{1}{l}{LJ}      & \multicolumn{1}{l}{CSG}              \\ \hline
\textbf{Qwen2-72B}                                       & 56.39\%                     & 52.03\%                              \\
\textbf{Qwen2.5-72B}                                     & 62.18\%                     & 54.51\%                              \\
\multicolumn{1}{l}{\textbf{Qwen2-72B vs Qwen2.5-72B}}    & 5.79\%                      & \textbf{2.48\%}                      \\ \hline
\textbf{Llama3-70B}                                      & 62.18\%                     & 59.55\%                              \\
\textbf{Llama3.1-70B}                                    & 53.70\%                     & 54.20\%                              \\
\textbf{LLama3-70B vs Llama3.1-70B}                      & 8.48\%                      & \textbf{5.35\%}                      \\ \hline
\multicolumn{1}{l}{\textbf{Qwen2.5-72B vs Llama3.1-70B}} & 8.48\%                      & \textbf{0.31\%}                      \\ \hline
\multicolumn{1}{l}{\textbf{Inconsistency with Human}}      & \multicolumn{1}{l}{24.35\%} & \multicolumn{1}{l}{\textbf{8.68\%}} \\ \hline
\end{tabular}
}
\end{table}

\setlength{\tabcolsep}{2.5pt}

\begin{table*}[h!]
\centering
\caption{Improvements to the retriever. Blue text indicates performance improvement. \Ori~represents the retrieval performance of fully answerable queries on the unoptimized retriever, \Bef~denotes performance on the unoptimized retriever when data has partial clue completeness. As shown, \Bef~ underperforms compared to \Ori, indicating that the existing retriever struggles with domain-specific corpora. After optimization, all retrievers achieve consistent performance improvements across all datasets.}
\label{tab:Dense-Retriever-Improvement}
\renewcommand{\arraystretch}{1.1}
\scalebox{0.66}{
\begin{tabular}{crccclcccl|ccclcccl}
\hline
\multicolumn{2}{c}{\multirow{3}{*}{\textbf{Precision@3 (\% / 100)}}} & \multicolumn{8}{c|}{\textbf{Games}} & \multicolumn{8}{c}{\textbf{Mediguides}} \\ \cline{3-18} 
\multicolumn{2}{c}{} & \multicolumn{4}{c|}{\textbf{Hearthstone}} & \multicolumn{4}{c|}{\textbf{Zelda}} & \multicolumn{4}{c|}{\textbf{Drug}} & \multicolumn{4}{c}{\textbf{Mayoclinic}} \\ \cline{3-18} 
\multicolumn{2}{c}{} & \multicolumn{1}{c|}{\textbf{\SmallOri}} & \textbf{\SmallBef} & \multicolumn{1}{c|}{\textbf{\SmallAft}} & \multicolumn{1}{l|}{\BoldBlueUparrow~(\%)} & \multicolumn{1}{c|}{\textbf{\SmallOri}} & \textbf{\SmallBef} & \multicolumn{1}{c|}{\textbf{\SmallAft}} & \BoldBlueUparrow~(\%) & \multicolumn{1}{c|}{\textbf{\SmallOri}} & \textbf{\SmallBef} & \multicolumn{1}{c|}{\textbf{\SmallAft}} & \multicolumn{1}{l|}{\BoldBlueUparrow~(\%)} & \multicolumn{1}{c|}{\textbf{\SmallOri}} & \textbf{\SmallBef} & \multicolumn{1}{c|}{\textbf{\SmallAft}} & \BoldBlueUparrow~(\%) \\ \hline
\multirow{2}{*}{\textbf{\begin{tabular}[c]{@{}c@{}}stella\_en\_400M\_v5\\\citeStella\end{tabular}}} & \SH & \multicolumn{1}{c|}{\graytext{26.78}} & 20.94 & \multicolumn{1}{c|}{32.22} & \multicolumn{1}{l|}{\BoldBlueUparrow~\bluetext{53.87}} & \multicolumn{1}{c|}{\graytext{34.48}} & 30.68 & \multicolumn{1}{c|}{49.88} & \BoldBlueUparrow~\bluetext{62.58} & \multicolumn{1}{c|}{\graytext{69.49}} & 62.61 & \multicolumn{1}{c|}{72.24} & \multicolumn{1}{l|}{\BoldBlueUparrow~\bluetext{15.38}} & \multicolumn{1}{c|}{\graytext{49.07}} & 49.45 & \multicolumn{1}{c|}{60.17} & \BoldBlueUparrow~\bluetext{21.68} \\
 & \MH & \multicolumn{1}{c|}{\graytext{13.55}} & 12.38 & \multicolumn{1}{c|}{23.98} & \multicolumn{1}{l|}{\BoldBlueUparrow~\bluetext{93.70}} & \multicolumn{1}{c|}{\graytext{11.11}} & 8.33 & \multicolumn{1}{c|}{25.97} & \BoldBlueUparrow~\bluetext{211.8} & \multicolumn{1}{c|}{\graytext{32.97}} & 22.09 & \multicolumn{1}{c|}{46.51} & \multicolumn{1}{l|}{\BoldBlueUparrow~\bluetext{110.6}} & \multicolumn{1}{c|}{\graytext{31.98}} & 30.33 & \multicolumn{1}{c|}{37.71} & \BoldBlueUparrow~\bluetext{24.33} \\ \hline
\multirow{2}{*}{\textbf{\begin{tabular}[c]{@{}c@{}}gte-base-multilingual\\\citeGTE\end{tabular}}} & \SH & \multicolumn{1}{c|}{\graytext{34.02}} & 25.15 & \multicolumn{1}{c|}{41.26} & \multicolumn{1}{l|}{\BoldBlueUparrow~\bluetext{64.06}} & \multicolumn{1}{c|}{\graytext{43.18}} & 36.55 & \multicolumn{1}{c|}{51.96} & \BoldBlueUparrow~\bluetext{42.16} & \multicolumn{1}{c|}{\graytext{65.38}} & 50.70 & \multicolumn{1}{c|}{71.48} & \multicolumn{1}{l|}{\BoldBlueUparrow~\bluetext{40.99}} & \multicolumn{1}{c|}{\graytext{56.15}} & 56.03 & \multicolumn{1}{c|}{68.33} & \BoldBlueUparrow~\bluetext{21.95} \\
 & \MH & \multicolumn{1}{c|}{\graytext{24.66}} & 23.73 & \multicolumn{1}{c|}{33.67} & \multicolumn{1}{l|}{\BoldBlueUparrow~\bluetext{41.89}} & \multicolumn{1}{c|}{\graytext{20.27}} & 22.34 & \multicolumn{1}{c|}{30.56} & \BoldBlueUparrow~\bluetext{36.79} & \multicolumn{1}{c|}{\graytext{36.98}} & 18.22 & \multicolumn{1}{c|}{53.35} & \multicolumn{1}{l|}{\BoldBlueUparrow~\bluetext{192.8}} & \multicolumn{1}{c|}{\graytext{37.32}} & 34.83 & \multicolumn{1}{c|}{46.62} & \BoldBlueUparrow~\bluetext{33.85} \\ \hline
\multirow{2}{*}{\textbf{\begin{tabular}[c]{@{}c@{}}snowflake-arctic-m-long\\\citeSnowflake\end{tabular}}} & \SH & \multicolumn{1}{c|}{\graytext{8.74}} & 9.54 & \multicolumn{1}{c|}{36.43} & \multicolumn{1}{l|}{\BoldBlueUparrow~\bluetext{281.9}} & \multicolumn{1}{c|}{\graytext{13.29}} & 15.65 & \multicolumn{1}{c|}{44.58} & \BoldBlueUparrow~\bluetext{184.9} & \multicolumn{1}{c|}{\graytext{32.32}} & 18.88 & \multicolumn{1}{c|}{75.51} & \multicolumn{1}{l|}{\BoldBlueUparrow~\bluetext{300.0}} & \multicolumn{1}{c|}{\graytext{24.59}} & 20.58 & \multicolumn{1}{c|}{68.57} & \BoldBlueUparrow~\bluetext{233.2} \\
 & \MH & \multicolumn{1}{c|}{\graytext{7.31}} & 9.36 & \multicolumn{1}{c|}{36.50} & \multicolumn{1}{l|}{\BoldBlueUparrow~\bluetext{290.0}} & \multicolumn{1}{c|}{\graytext{6.58}} & 9.42 & \multicolumn{1}{c|}{26.60} & \BoldBlueUparrow~\bluetext{182.4} & \multicolumn{1}{c|}{\graytext{15.03}} & 10.17 & \multicolumn{1}{c|}{56.22} & \multicolumn{1}{l|}{\BoldBlueUparrow~\bluetext{452.8}} & \multicolumn{1}{c|}{\graytext{16.10}} & 13.89 & \multicolumn{1}{c|}{48.57} & \BoldBlueUparrow~\bluetext{249.7} \\ \hline
\multirow{2}{*}{\textbf{\begin{tabular}[c]{@{}c@{}}snowflake-arctic-v1.5\\\citeSnowflake\end{tabular}}} & \SH & \multicolumn{1}{c|}{\graytext{9.08}} & 12.39 & \multicolumn{1}{c|}{33.21} & \multicolumn{1}{l|}{\BoldBlueUparrow~\bluetext{168.0}} & \multicolumn{1}{c|}{\graytext{11.80}} & 19.56 & \multicolumn{1}{c|}{46.09} & \BoldBlueUparrow~\bluetext{135.6} & \multicolumn{1}{c|}{\graytext{43.95}} & 54.12 & \multicolumn{1}{c|}{70.34} & \multicolumn{1}{l|}{\BoldBlueUparrow~\bluetext{29.97}} & \multicolumn{1}{c|}{\graytext{21.11}} & 36.18 & \multicolumn{1}{c|}{66.99} & \BoldBlueUparrow~\bluetext{85.16} \\
 & \MH & \multicolumn{1}{c|}{\graytext{9.32}} & 14.67 & \multicolumn{1}{c|}{36.21} & \multicolumn{1}{l|}{\BoldBlueUparrow~\bluetext{146.8}} & \multicolumn{1}{c|}{\graytext{4.53}} & 9.42 & \multicolumn{1}{c|}{31.52} & \BoldBlueUparrow~\bluetext{234.6} & \multicolumn{1}{c|}{\graytext{35.97}} & 39.05 & \multicolumn{1}{c|}{45.54} & \multicolumn{1}{l|}{\BoldBlueUparrow~\bluetext{16.62}} & \multicolumn{1}{c|}{\graytext{25.36}} & 31.98 & \multicolumn{1}{c|}{47.82} & \BoldBlueUparrow~\bluetext{49.53} \\ \hline
\multirow{2}{*}{\textbf{\begin{tabular}[c]{@{}c@{}}rubert-tiny-turbo\\\citeRubert\end{tabular}}} & \SH & \multicolumn{1}{c|}{\graytext{0.23}} & 0.00 & \multicolumn{1}{c|}{9.17} & \multicolumn{1}{l|}{\BoldBlueUparrow~\bluetext{$\infty$}} & \multicolumn{1}{c|}{\graytext{2.41}} & 1.22 & \multicolumn{1}{c|}{10.64} & \BoldBlueUparrow~\bluetext{772.1} & \multicolumn{1}{c|}{\graytext{7.87}} & 3.30 & \multicolumn{1}{c|}{19.26} & \multicolumn{1}{l|}{\BoldBlueUparrow~\bluetext{483.6}} & \multicolumn{1}{c|}{\graytext{12.99}} & 9.01 & \multicolumn{1}{c|}{22.53} & \BoldBlueUparrow~\bluetext{150.1} \\
 & \MH & \multicolumn{1}{c|}{\graytext{0.00}} & 0.00 & \multicolumn{1}{c|}{8.67} & \multicolumn{1}{l|}{\BoldBlueUparrow~\bluetext{$\infty$}} & \multicolumn{1}{c|}{\graytext{0.00}} & 0.00 & \multicolumn{1}{c|}{8.57} & \BoldBlueUparrow~\bluetext{$\infty$} & \multicolumn{1}{c|}{\graytext{1.82}} & 1.65 & \multicolumn{1}{c|}{13.18} & \multicolumn{1}{l|}{\BoldBlueUparrow~\bluetext{698.8}} & \multicolumn{1}{c|}{\graytext{10.68}} & 8.11 & \multicolumn{1}{c|}{21.65} & \BoldBlueUparrow~\bluetext{167.0} \\ \hline
\multirow{2}{*}{\textbf{\begin{tabular}[c]{@{}c@{}}MedEmbed-small-v0.1\\\citeMedEmbed\end{tabular}}} & \SH & \multicolumn{1}{c|}{\graytext{28.05}} & 20.94 & \multicolumn{1}{c|}{29.62} & \multicolumn{1}{l|}{\BoldBlueUparrow~\bluetext{41.45}} & \multicolumn{1}{c|}{\graytext{37.92}} & 27.75 & \multicolumn{1}{c|}{43.30} & \BoldBlueUparrow~\bluetext{56.04} & \multicolumn{1}{c|}{\graytext{65.74}} & 55.51 & \multicolumn{1}{c|}{71.61} & \multicolumn{1}{l|}{\BoldBlueUparrow~\bluetext{29.00}} & \multicolumn{1}{c|}{\graytext{56.84}} & 53.59 & \multicolumn{1}{c|}{63.82} & \BoldBlueUparrow~\bluetext{19.09} \\
 & \MH & \multicolumn{1}{c|}{\graytext{26.28}} & 23.73 & \multicolumn{1}{c|}{32.12} & \multicolumn{1}{l|}{\BoldBlueUparrow~\bluetext{35.36}} & \multicolumn{1}{c|}{\graytext{19.55}} & 14.86 & \multicolumn{1}{c|}{31.40} & \BoldBlueUparrow~\bluetext{111.3} & \multicolumn{1}{c|}{\graytext{46.90}} & 35.85 & \multicolumn{1}{c|}{54.77} & \multicolumn{1}{l|}{\BoldBlueUparrow~\bluetext{52.78}} & \multicolumn{1}{c|}{\graytext{39.32}} & 38.14 & \multicolumn{1}{c|}{47.98} & \BoldBlueUparrow~\bluetext{25.80} \\ \hline
\end{tabular}
}
\end{table*}

\subsection{Task Metrics}
\label{sec:TaskMetrics}

\textbf{Retrieval} To align with the RAG system's requirements, we set each retrieval to return \( k \in \{3\} \) chunks. The RAG system can be iteratively retrieved multiple times to gather the information deemed necessary by the LLM. Therefore, we assess retrieval using \( \text{Precision@k}^{\text{avg}} = \frac{1}{m} \left( \sum_{i=1}^m \frac{|\{D \in \mathcal{D}^{k}(q_{i}) \mid D \in \hat{\mathcal{D}}(q_{i})\}|}{k} \right) \), and \(\mathcal{D}^k(q_{i})\) denotes the top \( k \) documents retrieved by the retriever.

\textbf{Entire RAG system} 
To evaluate the overall RAG system, existing metrics include the Exact Match (EM) and LLM-as-a-Judge (LJ)~\cite{gu2025surveyllmasajudge}. The EM can provide stable evaluations but lacks robustness for complex responses and is sensitive to formatting variations. The LJ leverages LLM to assess the alignment between generated and reference answers, thereby mitigating some limitations of EM, but it can suffer from evaluation instability. To address these limitations, we introduce the \textit{Criteria-based Score for Generation} (CSG), \( s\equiv_{\mathrm{LLM}}\hat{\mathscr{A}}_i \) means the LLM confirms that \(s\) matches \( \hat{\mathscr{A}}_i \):

\begin{equation}
    \mathrm{CSG} = \frac{1}{m}\sum_{i=1}^m 
    \frac{\bigl|\{\,s\in\mathcal{S}(q_i)\mid s\equiv_{\mathrm{LLM}}\hat{\mathscr{A}}_i\}\bigr|}
    {|w_i|}\,.
\end{equation}

In the formula, $m$ is the number of queries; for each $q_i$, $\mathcal{S}(q_i)$ is the set of supporting sentences derived via $\mathscr{M}_1$ and $\mathscr{M}_2$, and $|w_i| \equiv |\mathcal{S}(q_i)|$.
The predicate $s \equiv_{\mathrm{LLM}} \hat{\mathscr{A}}_i$ denotes the LLM's criterion-aligned judgment that sentence $s$ satisfies the per-question rubric associated with $\hat{\mathscr{A}}_i$.
We provide complete variable definitions in Appendix~\ref{app:csg-defs}.

From Table~\ref{tab:the_superiority_of_CSG_scores}, we observe that the consistency of CSG is superior. This is primarily because the detailed criteria customized for each question in CSG can mitigate the problem of instability associated with using LLM-as-a-Judge alone. Similar conclusions have also been reached by other work, such as ~\cite{wei2025rocketeval} and ~\cite{kim-etal-2025-biggen}.
\section{Evaluation}
\label{sec:Evaluation}

\subsection{Setup}
\label{sec:Setup}

We use \texttt{Qwen2.5-72B-Instruct}. When generating data, we randomly sampled a \emph{subset} of corpus documents for synthesis (see Table~\ref{tab:Doc-Details}).
Empirically, this limited synthesis was sufficient to yield substantial gains in domain-specific retrieval, as evidenced by our main and ablation results.
A preliminary breakdown of resource usage is provided in Appendix~\ref{app:cost}.
Additional details of the synthesized training data are summarized in Appendix~\ref{app:details-synth-train-data}.

\textbf{Main Experiments}: The main experiment involves fine-tuning the retriever with synthetic data to enhance the performance and robustness of the retriever.

Based on the MTeb leaderboard~\cite{muennighoff-etal-2023-mteb}, we chose $6$ models of varying sizes and contexts that exhibited the best retrieval performance. The selected retrieval models, with their size and context length (CL): \texttt{stella\allowbreak \_\allowbreak en\allowbreak \_\allowbreak 400M\allowbreak \_\allowbreak v5} \cite{zhang2025jasperstelladistillationsota} (435M, CL 8192), \texttt{gte\allowbreak-multilingual\allowbreak-base} \cite{gte_multilingual_base} (611MB, CL 8192), \texttt{snowflake\allowbreak-arctic\allowbreak-embed\allowbreak-m\allowbreak-long/v1.5} \cite{merrick2024embeddingclusteringdataimprove}, (547M, CL 2048; 109M, CL 512), \texttt{rubert\allowbreak-tiny\allowbreak-turbo} \cite{rubert_tiny_turbo} (117M, CL 2048), and \texttt{MedEmbed\allowbreak-\allowbreak small\allowbreak -\allowbreak v0.1} \cite{medembed} (33M, CL 512). We optimize the retriever using contrastive loss~\cite{izacard2022unsupervised} and ANCE sampling~\cite{xiong2021approximate}. For convenience, we used the same hyperparameters across all experiments. Due to limitations in computational resources, we used LoRA with a relatively big rank to train uniformly for 10,000 steps with a learning rate of 2e-5. For detailed training parameters, please refer to the Appendix~\ref{app:detailed_training_parameters}.

\textbf{Generalization Experiments}: We evaluated the generalization performance of the trained models on the remaining 4 corpora.

\textbf{Overall RAG System Improvement Experiments:} For the entire RAG system, users care whether it can effectively solve their queries. We believe that optimizing the retriever can enhance the overall performance. We select 4 baselines: Vanilla~\cite{10.5555/3524938.3525306,DBLP:conf/icml/BorgeaudMHCRM0L22}, RR (Retrieve and Rank)~\cite{he2022rethinkingretrievalfaithfullarge}, where 3 results are retained per retrieval and 5 results are kept for generation after aggregating; Flare Style RAG~\cite{jiang-etal-2023-active}, which initially generates a sentence using a planner, then uses this sentence for retrieval to verify its accuracy, iterating up to a maximum of 8 times; and ReACT~\cite{yao2023react}, where the planner independently decides when to retrieve, aggregate, and stop, with a maximum of 8 interactions. The base model in use remains \texttt{Qwen2.5-72B-Instruct}.

\subsection{Main Experiments}
\label{sec:Main-Experiments}

Table \ref{tab:Dense-Retriever-Improvement} demonstrates the performance improvements. It can be observed that existing retrievers perform poorly when faced with domain-specific queries. However, their performance can be significantly improved after optimizing with \MYNAME. For convenience, we used the same hyperparameters across all experiments. As a result, models may not achieve optimal performance, indicating room for improvement. Additionally, we can clearly see that some retrievers trained on general data perform very poorly. This further underscores the importance of an automated RAG data synthesis framework tailored to specific domains for enhancing retriever performance.

\setlength{\tabcolsep}{6pt}

\begin{table*}[ht!]
\centering
\caption{Through continuous transformation of original questions, additional improvements can be achieved beyond those gained from domain-specific fine-tuning.}
\label{tab:Ablation}
\renewcommand{\arraystretch}{1.1}
\scalebox{0.76}{
\begin{tabular}{crcccc|cccc}
\hline
\multicolumn{2}{c}{\multirow{3}{*}{\textbf{Precision@3 (\% / 100)}}} & \multicolumn{4}{c|}{\textbf{Games}} & \multicolumn{4}{c}{\textbf{Mediguides}} \\ \cline{3-10} 
\multicolumn{2}{c}{} & \multicolumn{2}{c|}{\textbf{Hearthstone}} & \multicolumn{2}{c|}{\textbf{Zelda}} & \multicolumn{2}{c|}{\textbf{Drug}} & \multicolumn{2}{c}{\textbf{Mayoclinic}} \\ \cline{3-10} 
\multicolumn{2}{c}{} & \SH & \multicolumn{1}{c|}{\MH} & \SH & \MH & \SH & \multicolumn{1}{c|}{\MH} & \SH & \MH \\ \hline
\multirow{3}{*}{\textbf{\begin{tabular}[c]{@{}c@{}}stella\_en\_400M\_v5\\ \citeStella\end{tabular}}} & \textbf{w/o Logical Transformation} & 28.32 & \multicolumn{1}{c|}{20.18} & 45.23 & 22.24 & 65.23 & \multicolumn{1}{c|}{41.98} & 55.32 & 30.21 \\
 & \textbf{w/o Compleness Transformation} & 25.54 & \multicolumn{1}{c|}{18.23} & 43.33 & 21.45 & 64.32 & \multicolumn{1}{c|}{39.32} & 52.76 & 32.22 \\ \cline{2-10} 
 & \textbf{Full} & \bluetext{32.22} & \multicolumn{1}{c|}{\bluetext{23.98}} & \bluetext{49.88} & \bluetext{25.97} & \bluetext{72.24} & \multicolumn{1}{c|}{\bluetext{46.51}} & \bluetext{60.17} & \bluetext{37.71} \\ \hline
\multirow{3}{*}{\textbf{\begin{tabular}[c]{@{}c@{}}snowflake-arctic-v1.5\\ \citeSnowflake\end{tabular}}} & \textbf{w/o Logical Transformation} & 28.28 & \multicolumn{1}{c|}{34.32} & 43.78 & 28.75 & 64.32 & \multicolumn{1}{c|}{38.32} & 60.21 & 44.87 \\
 & \textbf{w/o Compleness Transformation} & 26.21 & \multicolumn{1}{c|}{30.21} & 41.76 & 25.69 & 62.23 & \multicolumn{1}{c|}{40.22} & 57.43 & 43.76 \\ \cline{2-10} 
 & \textbf{Full} & \bluetext{33.21} & \multicolumn{1}{c|}{\bluetext{36.21}} & \bluetext{46.09} & \bluetext{31.52} & \bluetext{70.34} & \multicolumn{1}{c|}{\bluetext{45.54}} & \bluetext{66.94} & \bluetext{47.82} \\ \hline
\multirow{3}{*}{\textbf{\begin{tabular}[c]{@{}c@{}}MedEmbed-small-v0.1\\ \citeMedEmbed\end{tabular}}} & \textbf{w/o Logical Transformation} & 28.32 & \multicolumn{1}{c|}{\bluetext{32.56}} & 41.23 & 26.43 & 63.42 & \multicolumn{1}{c|}{47.32} & 58.34 & 42.23 \\
 & \textbf{w/o Compleness Transformation} & 25.38 & \multicolumn{1}{c|}{30.34} & 37.24 & 24.24 & 62.56 & \multicolumn{1}{c|}{50.35} & 53.43 & 45.34 \\ \cline{2-10} 
 & \textbf{Full} & \bluetext{29.62} & \multicolumn{1}{c|}{32.12} & \bluetext{43.30} & \bluetext{31.40} & \bluetext{71.61} & \multicolumn{1}{c|}{\bluetext{54.77}} & \bluetext{63.82} & \bluetext{47.98} \\ \hline
\end{tabular}
}
\end{table*}

\setlength{\tabcolsep}{1.5pt}

\begin{table*}[ht!]
\centering
\caption{The retriever fine-tuned on Zelda consistently enhances the retrievers on other domains.}
\label{tab:Retriever-Generalization}
\renewcommand{\arraystretch}{1.18}
\scalebox{0.66}{
\begin{tabular}{clcccc|cccc|cc|cccc}
\hline
 &  & \multicolumn{4}{c|}{\textbf{Universities}} & \multicolumn{4}{c|}{\textbf{Documentation}} & \multicolumn{2}{c|}{\textbf{Games}} & \multicolumn{4}{c}{\textbf{Mediguides}} \\ \cline{2-16} 
 &  & \multicolumn{2}{c|}{\textbf{Stanford}} & \multicolumn{2}{c|}{\textbf{Berkeley}} & \multicolumn{2}{c|}{\textbf{Cyotek}} & \multicolumn{2}{c|}{\textbf{Notion}} & \multicolumn{2}{c|}{\textbf{Hearthstone}} & \multicolumn{2}{c}{\textbf{Drug}} & \multicolumn{2}{c}{\textbf{Mayoclinic}} \\ \cline{2-16} 
\multirow{-3}{*}{\textbf{Precision@3 (\% / 100)}} &  & \SH & \multicolumn{1}{c|}{\MH} & \SH & \MH & \SH & \multicolumn{1}{c|}{\MH} & \SH & \MH & \SH & \MH & \SH & \MH & \SH & \MH \\ \hline
 & \textbf{\SmallBef} & 59.52 & \multicolumn{1}{c|}{27.05} & 37.96 & 15.79 & 47.80 & \multicolumn{1}{c|}{25.72} & 13.77 & 2.21 & 20.94 & 12.38 & 62.61 & 22.09 & 49.45 & 30.33 \\
 & \textbf{\SmallAft} & 61.90 & \multicolumn{1}{c|}{31.72} & 39.19 & 16.52 & 51.22 & \multicolumn{1}{c|}{29.31} & 14.49 & 3.39 & 25.40 & 18.11 & 63.24 & 25.48 & 52.98 & 32.13 \\ \cline{2-16} 
\multirow{-3}{*}{\textbf{\begin{tabular}[c]{@{}c@{}}stella\_en\_400M\_v5\\\citeStella\end{tabular}}} & \textbf{\BoldBlueUparrow~(\%)} & \BoldBlueUparrow~\bluetext{3.40} & \multicolumn{1}{c|}{\BoldBlueUparrow~\bluetext{14.72}} & \BoldBlueUparrow~\bluetext{3.24} & \BoldBlueUparrow~\bluetext{4.62} & \BoldBlueUparrow~\bluetext{7.15} & \multicolumn{1}{c|}{\BoldBlueUparrow~\bluetext{13.96}} & \BoldBlueUparrow~\bluetext{5.23} & \BoldBlueUparrow~\bluetext{53.39} & \BoldBlueUparrow~\bluetext{21.30} & \BoldBlueUparrow~\bluetext{46.28} & \BoldBlueUparrow~\bluetext{1.01} & \BoldBlueUparrow~\bluetext{15.35} & \BoldBlueUparrow~\bluetext{7.14} & \BoldBlueUparrow~\bluetext{5.93} \\ \hline
 & \textbf{\SmallBef} & 48.41 & \multicolumn{1}{c|}{24.77} & 26.04 & 10.22 & 40.33 & \multicolumn{1}{c|}{26.44} & 11.74 & 3.98 & 12.39 & 14.67 & 54.12 & 39.05 & 36.18 & 31.98 \\
 & \textbf{\SmallAft} & 59.13 & \multicolumn{1}{c|}{28.54} & 39.93 & 13.20 & 51.38 & \multicolumn{1}{c|}{30.53} & 18.20 & 4.26 & 24.41 & 24.61 & 66.79 & 44.96 & 56.64 & 39.79 \\ \cline{2-16} 
\multirow{-3}{*}{\textbf{\begin{tabular}[c]{@{}c@{}}snowflake-arctic-v1.5\\\citeSnowflake\end{tabular}}} & \textbf{\BoldBlueUparrow~(\%)} & \BoldBlueUparrow~\bluetext{22.14} & \multicolumn{1}{c|}{\BoldBlueUparrow~\bluetext{15.22}} & \BoldBlueUparrow~\bluetext{53.34} & \BoldBlueUparrow~\bluetext{29.16} & \BoldBlueUparrow~\bluetext{27.40} & \multicolumn{1}{c|}{\BoldBlueUparrow~\bluetext{15.47}} & \BoldBlueUparrow~\bluetext{55.03} & \BoldBlueUparrow~\bluetext{7.04} & \BoldBlueUparrow~\bluetext{97.01} & \BoldBlueUparrow~\bluetext{67.76} & \BoldBlueUparrow~\bluetext{23.41} & \BoldBlueUparrow~\bluetext{15.13} & \BoldBlueUparrow~\bluetext{56.55} & \BoldBlueUparrow~\bluetext{24.42} \\ \hline
 & \textbf{\SmallBef} & 55.56 & \multicolumn{1}{c|}{23.40} & 35.26 & 12.81 & 43.90 & \multicolumn{1}{c|}{25.22} & 12.69 & 2.36 & 20.94 & 23.73 & 55.51 & 35.85 & 53.59 & 38.14 \\
 & \textbf{\SmallAft} & 58.73 & \multicolumn{1}{c|}{28.77} & 37.97 & 14.80 & 48.13 & \multicolumn{1}{c|}{32.18} & 15.69 & 3.78 & 29.24 & 28.51 & 67.17 & 43.80 & 57.13 & 41.22 \\ \cline{2-16} 
\multirow{-3}{*}{\textbf{\begin{tabular}[c]{@{}c@{}}MedEmbed-small-v0.1\\\citeMedEmbed\end{tabular}}} & \textbf{\BoldBlueUparrow~(\%)} & \BoldBlueUparrow~\bluetext{5.71} & \multicolumn{1}{c|}{\BoldBlueUparrow~\bluetext{22.95}} & \BoldBlueUparrow~\bluetext{7.69} & \BoldBlueUparrow~\bluetext{15.53} & \BoldBlueUparrow~\bluetext{9.64} & \multicolumn{1}{c|}{\BoldBlueUparrow~\bluetext{27.60}} & \BoldBlueUparrow~\bluetext{23.64} & \BoldBlueUparrow~\bluetext{60.17} & \BoldBlueUparrow~\bluetext{39.64} & \BoldBlueUparrow~\bluetext{20.14} & \BoldBlueUparrow~\bluetext{21.01} & \BoldBlueUparrow~\bluetext{22.18} & \BoldBlueUparrow~\bluetext{6.61} & \BoldBlueUparrow~\bluetext{8.08} \\ \hline
\end{tabular}
}
\end{table*}

\subsection{Ablation Experiments}
\label{sec:Ablation-Experiments}

Table~\ref{tab:Ablation} presents the performance of our various data augmentation methods, specifically logical transformations and completeness transformations, across different domain datasets. It is evident that the new data synthesized through logical and completeness transformations significantly enhances the retriever's effectiveness on domain-specific data. This approach facilitates a deeper understanding of domain-specific datasets by the retriever.

More ablation study results about the retriever are reported in Appendix~\ref{app:full-ablation}.

\subsection{Generalization Experiments}
\label{sec:Generalization-Experiments}

Tables \ref{tab:Retriever-Generalization} present the performance of retrievers when fine-tuned on Zelda and applied to other datasets. It is evident that the data fine-tuned with \MYNAME~generalizes well to out-of-domain tasks for retrievers. This implies that logical transformations and completeness transformations are not only effective for training in specific domains but also possess a certain degree of generalization capability for out-of-domain problems.

\setlength{\tabcolsep}{4pt}

\begin{table*}[ht!]
\centering
\caption{The \MYNAME-optimized retriever consistently enhances the performance of both single-hop and multi-hop RAG across various methods. This improvement primarily stems from the retriever's enhanced understanding of the fine-grained semantics in user queries, enabling the LLM to obtain more robust retrieval results with fewer interaction steps through more casual sub-queries. SH stands for Single Hop, and MH stands for Multi Hop.}
\label{tab:Overall-RAG-System-Experiments}
\renewcommand{\arraystretch}{1.15}
\scalebox{0.78}{
\begin{tabular}{clcccc|cc|cc|cc}
\hline
\multirow{3}{*}{\textbf{\begin{tabular}[c]{@{}c@{}}Precision@3 \\ (\% / 100)\end{tabular}}} &  & \multicolumn{4}{c|}{\textbf{\begin{tabular}[c]{@{}c@{}}ReACT Style\\ \citeReACT\end{tabular}}} & \multicolumn{2}{c|}{\textbf{\begin{tabular}[c]{@{}c@{}}vanillaRAG\\ \citevanillaRAG\end{tabular}}} & \multicolumn{2}{c|}{\textbf{\begin{tabular}[c]{@{}c@{}}Flare Style\\ \citeFlare\end{tabular}}} & \multicolumn{2}{c}{\textbf{\begin{tabular}[c]{@{}c@{}}RR Style\\ \citeRR\end{tabular}}} \\ \cline{2-12} 
 &  & \multicolumn{2}{c|}{\textbf{4-Iter}} & \multicolumn{2}{c|}{\textbf{8-Iter}} & \multicolumn{1}{c|}{\multirow{2}{*}{\SH}} & \multirow{2}{*}{\MH} & \multicolumn{1}{c|}{\multirow{2}{*}{\SH}} & \multirow{2}{*}{\MH} & \multicolumn{1}{c|}{\multirow{2}{*}{\SH}} & \multirow{2}{*}{\MH} \\ \cline{2-6}
 &  & \multicolumn{1}{l}{\SH} & \multicolumn{1}{l|}{\MH} & \multicolumn{1}{l}{\SH} & \multicolumn{1}{l|}{\MH} & \multicolumn{1}{c|}{} &  & \multicolumn{1}{c|}{} &  & \multicolumn{1}{c|}{} &  \\ \hline
\multirow{3}{*}{\textbf{\begin{tabular}[c]{@{}c@{}}stella\_en\_400M\_v5\\ \citeStella\end{tabular}}} & \textbf{\SmallBef} & 51.52 & \multicolumn{1}{c|}{39.43} & 39.05 & 53.22 & \multicolumn{1}{c|}{49.00} & 43.88 & \multicolumn{1}{c|}{51.86} & 57.40 & \multicolumn{1}{c|}{57.62} & 53.79 \\
 & \textbf{\SmallAft} & 52.45 & \multicolumn{1}{c|}{43.32} & 43.23 & 55.23 & \multicolumn{1}{c|}{53.24} & 45.72 & \multicolumn{1}{c|}{55.34} & 60.23 & \multicolumn{1}{c|}{62.34} & 56.43 \\ \cline{2-12} 
 & \textbf{\BoldBlueUparrow~(\%)} & \BoldBlueUparrow~\bluetext{1.81} & \multicolumn{1}{c|}{\BoldBlueUparrow~\bluetext{9.87}} & \BoldBlueUparrow~\bluetext{10.93} & \BoldBlueUparrow~\bluetext{3.78} & \multicolumn{1}{c|}{\BoldBlueUparrow~\bluetext{8.65}} & \BoldBlueUparrow~\bluetext{4.19} & \multicolumn{1}{c|}{\BoldBlueUparrow~\bluetext{6.71}} & \BoldBlueUparrow~\bluetext{4.93} & \multicolumn{1}{c|}{\BoldBlueUparrow~\bluetext{8.19}} & \BoldBlueUparrow~\bluetext{4.91} \\ \hline
\multirow{3}{*}{\textbf{\begin{tabular}[c]{@{}c@{}}snowflake-arctic-v1.5\\ \citeSnowflake\end{tabular}}} & \textbf{\SmallBef} & 45.19 & \multicolumn{1}{c|}{39.88} & 38.19 & 44.99 & \multicolumn{1}{c|}{37.05} & 34.76 & \multicolumn{1}{c|}{50.12} & 59.48 & \multicolumn{1}{c|}{53.40} & 50.57 \\
 & \textbf{\SmallAft} & 46.95 & \multicolumn{1}{c|}{42.38} & 42.19 & 47.17 & \multicolumn{1}{c|}{40.21} & 41.85 & \multicolumn{1}{c|}{55.23} & 58.14 & \multicolumn{1}{c|}{57.57} & 55.25 \\ \cline{2-12} 
 & \textbf{\BoldBlueUparrow~(\%)} & \BoldBlueUparrow~\bluetext{3.89} & \multicolumn{1}{c|}{\BoldBlueUparrow~\bluetext{6.27}} & \BoldBlueUparrow~\bluetext{10.47} & \BoldBlueUparrow~\bluetext{4.85} & \multicolumn{1}{c|}{\BoldBlueUparrow~\bluetext{8.53}} & \BoldBlueUparrow~\bluetext{20.40} & \multicolumn{1}{c|}{\BoldBlueUparrow~\bluetext{10.20}} & \BoldRedDownarrow~\redtext{2.25} & \multicolumn{1}{c|}{\BoldBlueUparrow~\bluetext{7.81}} & \BoldBlueUparrow~\bluetext{9.25} \\ \hline
\multirow{3}{*}{\textbf{\begin{tabular}[c]{@{}c@{}}MedEmbed-small-v0.1\\ \citeMedEmbed\end{tabular}}} & \textbf{\SmallBef} & 45.47 & \multicolumn{1}{c|}{40.41} & 44.92 & 39.94 & \multicolumn{1}{c|}{40.33} & 32.54 & \multicolumn{1}{c|}{46.80} & 58.43 & \multicolumn{1}{c|}{50.73} & 53.40 \\
 & \textbf{\SmallAft} & 45.66 & \multicolumn{1}{c|}{44.95} & 44.05 & 44.35 & \multicolumn{1}{c|}{43.16} & 41.62 & \multicolumn{1}{c|}{54.95} & 66.77 & \multicolumn{1}{c|}{54.11} & 52.84 \\ \cline{2-12} 
 & \textbf{\BoldBlueUparrow~(\%)} & \BoldBlueUparrow~\bluetext{0.42} & \multicolumn{1}{c|}{\BoldBlueUparrow~\bluetext{0.42}} & \BoldRedDownarrow~\redtext{1.94} & \BoldBlueUparrow~\bluetext{11.04} & \multicolumn{1}{c|}{\BoldBlueUparrow~\bluetext{7.02}} & \BoldBlueUparrow~\bluetext{27.90} & \multicolumn{1}{c|}{\BoldBlueUparrow~\bluetext{17.41}} & \BoldBlueUparrow~\bluetext{14.27} & \multicolumn{1}{c|}{\BoldBlueUparrow~\bluetext{6.63}} & \BoldRedDownarrow~\redtext{1.05} \\ \hline
\end{tabular}
}
\end{table*}

\subsection{Overall RAG System Experiments}
\label{sec:Overall-RAG-System-Experiments}

Tables \ref{tab:Overall-RAG-System-Experiments} show the performance improvements across the entire RAG system when using a fine-tuned retriever. It is evident that a robust retriever can lead to steady enhancements in the overall performance of the RAG system. This improvement may primarily stems from the enhanced robustness of the retriever to queries, achieved through training with completeness data, which develops robust adaptability to various queries and subqueries. Our retriever not only retrieves queries and documents with direct semantic correspondence more accurately but also demonstrates strong robustness in cases where queries and documents are partially misaligned.
\vspace{-0.5em}

\section{Conclusion}
\label{sec:Conclusion}

In this work, we introduced \MYNAME, a framework for synthesizing RAG data that captures the complex relationships between queries, documents, answers, and supporting clues. \MYNAME~enables the generation of diverse, domain-specific datasets with varying logical complexities and clue completeness, enhancing the robustness of retrievers. We also develop \MYBench, a benchmark spanning 8 document corpora across 4 domains. The extensive experiments demonstrated significant improvements in retriever performance across multiple domains and the RAG paradigm. These results validate the effectiveness and generalizability of \MYNAME.
\section{Limitations}
\label{sec:Limitations}

\textbf{The types of synthesis and evaluation questions are limited.} The existing implementation of \MYNAME~concentrates on generating fact-based QA data with deterministic answers, rather than tackling more complex questions that involve logical reasoning across different entities within documents. This issue is not unique to our work; current research also focuses on evaluating "factual questions" for ease of assessment. Further exploration of open-ended questions remains necessary.

\textbf{Assessing the Authenticity of Benchmark Questions.} We cannot guarantee that the questions we construct reflect those that would be genuinely asked by humans in the real world. We believe this is something that cannot be ensured for most existing RAG evaluations, as their QA datasets are rarely collected from real-world scenarios. Instead, they are often crowdsourced or generated by language models. However, we do not see this as entirely necessary, as we have conducted generalization experiments that demonstrate the model's ability to generalize to some extent.

\textbf{Limitations of Domain Selection.} Our research focuses on domain-specific retrieval optimization up to RAG system optimization. We have made efforts to collect publicly available domain-specific datasets that we consider valuable and have used LLMs for data cleansing. However, in the real world, domain-specific data is more complex, and we cannot cover all possible scenarios.

\bibliography{custom}

\appendix

    \section{Appendix}
    \label{sec:appendix}
    
    \subsection{Detailed Source Data Cleaning Procedures}
    \label{app:cleaning}
    
    The source data we obtained is from web pages. Therefore, we conducted the following data cleaning processes: extracting the main HTML content to eliminate irrelevant information, redirecting hyperlinks to convert them into local links, and converting files of different formats, such as extracting the main content from HTML and converting PDFs to text. Subsequently, we processed the resulting files using \texttt{Qwen2.5-72B-Instruct} to further remove any potentially invalid information from the original files. A cleaned sample prompt is as follows:
    
    \begin{tcolorbox}[
      enhanced,                   
      breakable,                  
      colback=gray!5,             
      colframe=blue!60!black,     
      boxrule=0.8pt,              
      arc=2mm,                    
      left=4pt, right=4pt,        
      top=4pt, bottom=4pt,
      width=\linewidth,           
      title={\bfseries System \& User Prompts},
      fonttitle=\bfseries
    ]
    {\ttfamily\small
    
    \textbf{System Prompt}
    
    You are an assistant tasked with cleaning and formatting text converted from HTML or PDF documents. The goal is to remove unnecessary content like formatting errors, irrelevant metadata, conversion artifacts, advertisements, logos, login pages, navigation menus, footers, social media buttons, copyright notices, terms and conditions, empty lists, incomplete sections, and other pages lacking useful content, while keeping the text true to the original meaning. Do not change the meaning of any part of the document. Ensure the output follows clean Markdown formatting.
    
    \vspace{1ex}
    \textbf{User Prompt}
    
    Your task is to:
    1. Remove all irrelevant or incomplete content mentioned.
    2. Correct any formatting issues and output the text in proper Markdown format.
    3. Keep the original meaning and content as accurate as possible.
    
    Here is the text:
    
    \{content\_chunk\}
    
    } 
    \end{tcolorbox}
    
    \subsection{Detailed Bench Construction Procedures and Examples}
    \label{app:bench}
    
    Our goal is to enhance retrieval and overall RAG system performance on domain-specific datasets. To achieve this, we need an evaluation benchmark to assess retrieval and RAG performance on these datasets. We established this benchmark using the following methods: (1) Single-hop Questions: After segmenting selected documents, each segment is randomly assigned to a domain-specific expert (e.g., Zelda enthusiasts or Hearthstone fans). These experts generate 3-6 questions per document segment using their perspective, assisted by LLMs or Google; (2) Multi-hop Questions: We leveraged the entity relationship graph component from \MYNAME~ to link multiple documents. Annotators were provided with documents connected by the same entity or relationship and tasked with generating 3-6 questions based on these documents. The corresponding answer sentences and answers were also recorded. All annotations are the author.
    
    The questions generated from these steps are typically direct Q\&A questions targeting entities or relationships. To simulate the variety of questions people might ask in real scenarios, we further enhanced each question by having different individuals modify them in various ways, such as introducing ambiguity, adding constraints, or including unanswerable yet related parts.
    
    \begin{tcolorbox}[
      enhanced,                   
      breakable,                  
      colback=gray!5,             
      colframe=blue!60!black,     
      boxrule=0.8pt,              
      arc=2mm,                    
      left=4pt, right=4pt,        
      top=4pt, bottom=4pt,
      width=\linewidth,           
      title={\bfseries Question Generation Instructions},
      fonttitle=\bfseries
    ]
    {\ttfamily\small
    
    1. Inputs  
    
      • Single‐hop: one document segment (~1–2 paragraphs on a single topic)
      
      • Multi‐hop: a small set (2–4) of segments linked by shared entities or relations  
    
    2. For each input, write 3–6 questions:  
    
      a. Single‐hop questions must be answerable by reading only that one segment.  
      
      b. Multi‐hop questions must require combining facts from all segments in the set.  
    
    3. Question style  
    
      • Focus on factual queries (Who/What/When/Where/How many/How).  
      
      • Keep each question clear and unambiguous.  
      
      • Do not introduce information not present in the segment(s).  
    
    4. Answer annotation  
    
      For each question, provide:  
      
      a. The exact sentence(s) from the segment(s) containing the answer (extractive span). 
      
      b. The concise answer text.  
    
    } 
    \end{tcolorbox}
    
    Finally, all generated questions were filtered using LLMs based on metrics like practical value, resulting in our final set of questions. Subsequently, we developed detailed scoring criteria for each problem, which were then verified manually.
    
    The compensation for generating each question is based on local pricing, that is 1 RMB per question in China, with additional payments for each enhanced variation.
    
    A sample question is as follows:
    
    \begin{tcolorbox}[
      enhanced,                   
      breakable,                  
      colback=gray!5,             
      colframe=blue!60!black,     
      boxrule=0.8pt,              
      arc=2mm,                    
      left=4pt, right=4pt,        
      top=4pt, bottom=4pt,
      width=\linewidth,           
      title={\bfseries A Question example},
      fonttitle=\bfseries
    ]
    {\ttfamily\small
    
    How many Side Quests are there in total in The Legend of Zelda: Breath of the Wild and its DLC Packs?
    
    } 
    \end{tcolorbox}

    A transformed sample is as follows:
    
    \begin{tcolorbox}[
      enhanced,                   
      breakable,                  
      colback=gray!5,             
      colframe=blue!60!black,     
      boxrule=0.8pt,              
      arc=2mm,                    
      left=4pt, right=4pt,        
      top=4pt, bottom=4pt,
      width=\linewidth,           
      title={\bfseries A Transformed Example},
      fonttitle=\bfseries
    ]
    {\ttfamily\small
    
    How many Side Quests were added to The Legend of Zelda: Breath of the Wild and its DLC Packs, and how does this number compare to the number of Side Quests in The Witcher 3: Wild Hunt?
    
    } 
    \end{tcolorbox}
    
    An example of a scoring criterion is as follows:
    
    \begin{tcolorbox}[
      enhanced,                   
      breakable,                  
      colback=gray!5,             
      colframe=blue!60!black,     
      boxrule=0.8pt,              
      arc=2mm,                    
      left=4pt, right=4pt,        
      top=4pt, bottom=4pt,
      width=\linewidth,           
      title={\bfseries Scoring Criterion Example},
      fonttitle=\bfseries
    ]
    {\ttfamily\small
    
    (1) Award 1 point if the response indicates that the respondent provides the correct quantity.
    (2) Award 1 point if the response indicates that the respondent is unaware of the number of side quests in *The Witcher 3: Wild Hunt*.
    
    } 
    \end{tcolorbox}

    \subsection{Details of the Synthetic Training Data}
    \label{app:details-synth-train-data}
    
    Considering the practical application value, we did not synthesize data from the entire corpus. Instead, we randomly sampled a subset of documents for data synthesis. The specific number of sampled documents is shown in the Table~\ref{tab:Doc-Details} under \texttt{\# Selected Docs for Synthesizing Data}. In the table, \texttt{\# Original Docs} refers to the number of documents obtained after cleaning the raw data. \texttt{\# Selected Docs as Corpus} indicates the number of documents used as the retrieval corpus, and \texttt{\# Selected Docs for Constructing Bench} refers to the documents used for building the evaluation QA set.
    
    
    \setlength{\tabcolsep}{2pt}
    
    \begin{table*}[h!]
    \centering
    \caption{Original document count, documents serving as the corpus (i.e., documents in our actual corpus), documents used for data synthesis, documents used for evaluation construction.}
    \vspace{-0.5em}
    \label{tab:Doc-Details}
    \renewcommand{\arraystretch}{1.25}
    \scalebox{0.8}{
    \begin{tabular}{ccc|cc|cc|cc}
    \hline
    \multirow{2}{*}{} & \multicolumn{2}{c|}{\textbf{Games}} & \multicolumn{2}{c|}{\textbf{Mediguides}} & \multicolumn{2}{c|}{\textbf{Universities}} & \multicolumn{2}{c}{\textbf{Documentation}} \\ \cline{2-9} 
     & \multicolumn{1}{c|}{\textbf{Hearthstone}} & \textbf{Zelda} & \multicolumn{1}{c|}{\textbf{Drug}} & \textbf{Mayoclinic} & \multicolumn{1}{c|}{\textbf{Stanford}} & \textbf{Berkeley} & \multicolumn{1}{c|}{\textbf{Cyotek}} & \textbf{Notion} \\ \hline
    \textbf{\# Original Docs} & \multicolumn{1}{c|}{10665} & 15695 & \multicolumn{1}{c|}{77348} & 2542 & \multicolumn{1}{c|}{65} & 280 & \multicolumn{1}{c|}{808} & 475 \\
    \textbf{\# Selected Docs as Corpus} & \multicolumn{1}{c|}{10665} & 15695 & \multicolumn{1}{c|}{24733} & 2542 & \multicolumn{1}{c|}{65} & 280 & \multicolumn{1}{c|}{808} & 475 \\
    \textbf{\# Selected Docs for Synthesizing Data} & \multicolumn{1}{c|}{870} & 873 & \multicolumn{1}{c|}{826} & 862 & \multicolumn{1}{c|}{/} & / & \multicolumn{1}{c|}{/} & / \\
    \textbf{\# Selected Docs for Constructing Bench} & \multicolumn{1}{c|}{100} & 100 & \multicolumn{1}{c|}{100} & 100 & \multicolumn{1}{c|}{65} & 100 & \multicolumn{1}{c|}{100} & 100 \\ \hline
    \end{tabular}
    }
    \end{table*}

    \subsection{Detailed Training Parameters}
    \label{app:detailed_training_parameters}
    
    Training settings are as follows: batch size of 32, 3 epochs, learning rate of 2e-5. All document indices are updated every 5 steps. The loss margin is set to 0.3. For LoRA, the rank is set to 128, alpha is set to 32, and dropout is set to 0.1.
    
    \subsection{Expert Annotator Qualifications}
    \label{app:experts}
    
    \paragraph{Games (Zelda, Hearthstone).}
    We recruited three enthusiasts: one knowledgeable in both domains and two domain-focused.
    For Zelda, two long-time Nintendo players each reported playing at least four mainline titles (average playtime $>$100 hours).
    For Hearthstone, two annotators reported average playtime $>$300 hours.
    
    \paragraph{Medical (Drug, Mayoclinic).}
    We recruited two graduate students in medical fields at a top-tier university.
    While not experts on every subtopic, both possess foundational biomedical knowledge and at least one focused area relevant to the corpora.
    
    \paragraph{Universities \& Documentation (Stanford, Berkeley, Cyotek, Notion).}
    These domains require no deep specialization; four student annotators per dataset were recruited based on prior familiarity (e.g., having browsed the relevant websites or used the software).
    
    \paragraph{Quality control.}
    All annotators followed question-generation and verification rubrics; two rounds of review were conducted for scoring rubrics prior to inclusion in the benchmark.
    
    \subsection{CSG Definitions}
    \label{app:csg-defs}
    
    We list key symbols used in CSG.
    For query $q_i$, let $\mathscr{A}_i$ denote the reference answer, $\mathcal{S}(q_i)$ the set of supporting sentences mapped via $\mathscr{M}_1$ (answer-to-clue) and $\mathscr{M}_2$ (clue-to-document/sentence), and $w_i \equiv |\mathcal{S}(q_i)|$.
    The predicate $s \equiv_{\mathrm{LLM}} \hat{\mathscr{A}}_i$ means the LLM, given the per-question rubric, confirms that sentence $s$ satisfies the criterion linked to $\hat{\mathscr{A}}_i$.
    
    \subsection{Preliminary Computational Cost Analysis}
    \label{app:cost}
    
    \paragraph{Synthesis scope.}
    We intentionally synthesize from a subset of documents per corpus (see Table~\ref{tab:Doc-Details}), which reduces data-generation overhead while remaining effective empirically.
    
    \paragraph{API/LMM costs.}
    Across all syntheses and filtering steps, the total API expenditure was approximately 200~USD.
    
    \subsection{Full Ablation Results Across All Retrievers}
    \label{app:full-ablation}
    
    This section reports the extended ablations (``w/o logical transformation'' and ``w/o completeness transformation'') for all retrievers evaluated in the main experiments, across all domains. Please refer to Table~\ref{tab:full-ablation-all-retrievers} for details.
    
    \begin{table*}[ht!]
    \centering
    \caption{Extended ablation across all evaluated retrievers (Precision@3 $\uparrow$). Columns correspond to the 8 domain-specific datasets: Zelda, Hearthstone, Drug, Mayo, Stanford, Berkeley, Cyotek, Notion. ``Full (ours)'' uses both logical and completeness transformations; the two ablations remove one transformation at a time.}
    \label{tab:full-ablation-all-retrievers}
    \renewcommand{\arraystretch}{1.25}
    \scalebox{0.88}{
    \begin{tabular}{lcccccccc}
    \toprule
    Method & Zelda & Hearthstone & Drug & Mayo & Stanford & Berkeley & Cyotek & Notion \\
    \midrule
    \multicolumn{9}{l}{\textbf{gte-base-multilingual}} \\
    \quad w/o Logical Transformation & 40.03 & 29.47 & 49.53 & 30.42 & 67.48 & 50.38 & 66.23 & 43.56 \\
    \quad w/o Completeness Transformation & 38.86 & 27.34 & 45.48 & 28.34 & 67.34 & 48.31 & 65.23 & 42.34 \\
    \quad \textbf{Full (ours)} & \textbf{41.26} & \textbf{33.67} & \textbf{51.96} & \textbf{30.56} & \textbf{71.48} & \textbf{53.35} & \textbf{68.33} & \textbf{46.62} \\
    \midrule
    \multicolumn{9}{l}{\textbf{snowflake-arctic-m-long}} \\
    \quad w/o Logical Transformation & 33.96 & 32.13 & 42.43 & 24.45 & 72.48 & 51.23 & 63.96 & 45.32 \\
    \quad w/o Completeness Transformation & 34.63 & 34.48 & 42.13 & 23.45 & 72.36 & 52.13 & 64.57 & 43.76 \\
    \quad \textbf{Full (ours)} & \textbf{36.43} & \textbf{36.50} & \textbf{44.58} & \textbf{26.60} & \textbf{75.51} & \textbf{56.22} & \textbf{68.57} & \textbf{48.57} \\
    \midrule
    \multicolumn{9}{l}{\textbf{rubert-tiny-turbo}} \\
    \quad w/o Logical Transformation & 9.10 & 7.23 & 9.32 & 8.12 & 17.31 & 11.07 & 20.13 & 19.45 \\
    \quad w/o Completeness Transformation & \textbf{9.29} & 8.23 & 9.47 & 8.21 & 16.13 & 11.18 & 17.23 & 18.54 \\
    \quad \textbf{Full (ours)} & 9.17 & \textbf{8.67} & \textbf{10.64} & \textbf{8.57} & \textbf{19.26} & \textbf{13.18} & \textbf{22.53} & \textbf{21.65} \\
    \bottomrule
    \end{tabular}
    }
    \end{table*}

\end{document}